# Control and Evaluation of a Humanoid Robot with Rolling Contact Knees

Seung Hyeon Bang[1], Carlos Gonzalez[1], Junhyeok Ahn[2], Nicholas Paine[3] and Luis Sentis[1]

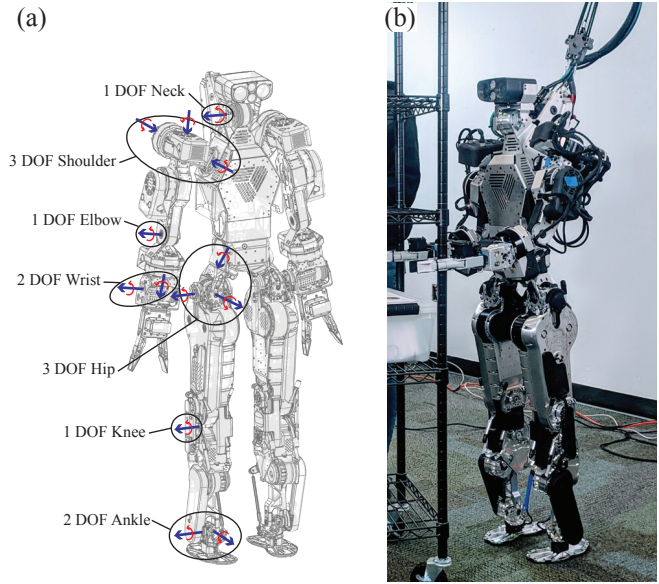

Fig. 1. **DRACO 3 Humanoid.** (a) Wireframe of DRACO 3 showing the degrees of freedom on the right side of the robot, and their respective axis of rotation. (b) DRACO 3 standing at full height.

*Abstract*— In this paper, we introduce the humanoid robot DRACO 3 by providing a high-level description of its design and control. This robot features proximal actuation and mechanical artifacts to provide a high range of hip, knee and ankle motion. Its versatile design brings interesting problems as it requires a more elaborate control system to perform its motions. For this reason, we introduce a whole body controller (WBC) with support for rolling contact joints and show how it can be easily integrated into our previously presented open-source Planning and Control (PnC) framework. We then validate our controller experimentally on DRACO 3 by showing preliminary results carrying out two postural tasks. Lastly, we analyze the impact of the proximal actuation design and show where it stands in comparison to other adult-size humanoids.

## I. INTRODUCTION

Dynamic motions for legged systems require accurate sensing of the robot's states and high performance control. To realize the latter, several important issues must be considered. On one hand, efficient transmissions with low friction/stiction and low backlash are desired. In addition, mechanical designs capable of achieving a wide range of motion (RoM) are important and non-trivial to achieve. On the other hand, these types of mechanisms come at the expense of an increase in complexity of the controllers needed to exploit the potential of the high-dimensional system.

Humanoids have often employed collocated actuation (motors directly located at each joint) because of its simplicity in design [1], [2]. However, the performance of these robots degrades when using simplified models for planning because of model discrepancy caused by the heavy distal mass on their legs [3]. Due to this, their design has shifted in favor of proximal actuation (placing heavy motors near the body) to reduce the limbs' distal mass for dynamic maneuvers. As part of the solution to achieve this, a transmission system that has been explored frequently in legged robots consists of cable-based drive systems [4]–[6]. This is due to their light weight as well as effective power transmission capability, thus helping provide high torque density when incorporated into the design. Additionally, this kind of transmission is more efficient because the mechanical losses due to friction are small and, thus, provides higher torque transparency.

In a similar manner, DRACO 3 (shown in Fig. 1) was designed bearing a cable-based drive system in mind for proximal actuation. In addition, a rolling contact joint (RCJ) mechanism is employed to enhance its RoM. Although RCJs have been widely used in other fields such as in lower extremity exoskeleton design to reduce misalignment between the human's and the device's joint [7], [8], and in robotic fingers to realize large RoM as well as to decrease internal friction [9], it has not been adopted on humanoids due to its mechanical complexity and complicated control design. By means of a RCJ on the knee, DRACO 3 achieved proximal actuation and high ranges of motion.

Typical methods to control complex systems such as legged robots often make use of a WBC, which compute joint-level commands to achieve desired operational space tasks. This type of controller is often designed by considering a model of the robot along with multiple physical and environmental constraints (e.g., friction). In addition, any complexities in the joint mechanisms must be taken into account either within or outside of the WBC module. The recent work by [10] presents a rich literature review on different types of WBCs. In this work, we chose to pursue a WBC, similar to that in [10], [11] for its high flexibility and ease of intuitive tuning. Previous works [7]–[9], [12] have studied the control of complex joint mechanisms (e.g.,

This work was supported by ONR Grant# N000141512507.

[1]S.H. Bang, C. Gonzalez, and L. Sentis are with the Department of Aerospace Engineering and Engineering Mechanics, The University of Texas at Austin, TX 78712, USA bangsh0718@utexas.edu, carlos.gonzalez@utexas.edu, lsentis@austin.utexas.edu

[2]J. Ahn is with the Department of Mechanical Engineering, The University of Texas at Austin, TX, 78712, USA junhyeokahn91@gmail.com

[3]N. Paine is with Apptronik, Inc., 110701 Stonehollow Dr STE 150, Austin, TX, 78758, USA n.a.paine@gmail.com

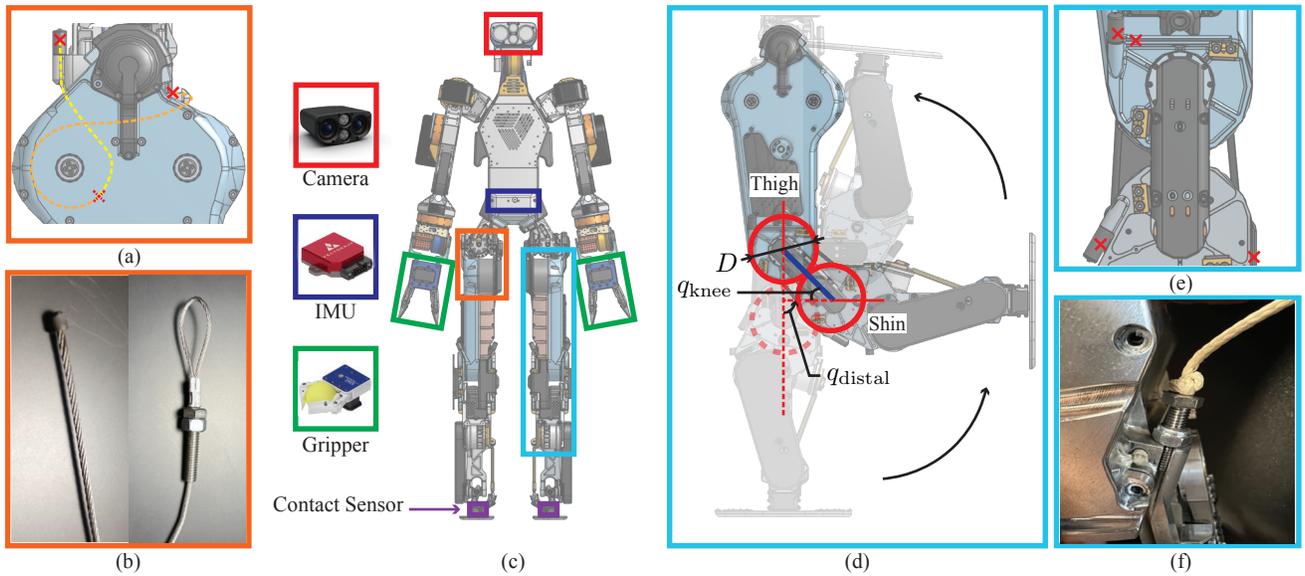

Fig. 2. **Mechatronic Components of DRACO 3. (a) Hip actuation**: Lateral view of thigh, showing cable routing (- -) and the location of the terminations (×) of the stainless steel cables. **(b) Cable terminations of the Hip pitch joint.** Soldered termination (left) and crimped termination (right). **(c) Mechatronic components in DRACO 3**: the camera, the IMU, the grippers, and the contact sensors are Carnegie Robotics MultiSense S7, VectorNav VN-100, SAKE Robotics EZGripper, Apptronik load cell, respectively. **(d) Rolling contact joints**. Snapshots of leg configurations in sequence when $q_{knee}$ is 0°, 90°, and 180°. **(e) Knee joint**: Lateral view of knee, showing the location of its cable terminations (×). **(f) Cable terminations of the Knee joint**.

rolling contact joints) in an isolated manner, by developing a high fidelity characterization for the modelled and controlled joint. The presence of such complex joint mechanisms in high-dimensional robotic systems is emerging and, similarly, control architectures and WBCs that perform effectively using these complex mechanisms. Some WBCs that use non-traditional collocated actuators include the use of parallel linear actuators, which can be solved through a transmission mapping [2], [13]. Others have taken into account in the WBC formulation both series and parallel mechanisms by defining their optimization problem in actuation space [14]. However, to the best of our knowledge, this is the first WBC that takes into account a RCJ mechanism and is successfully deployed on a high-dimensional legged robot.

The main contributions of this paper are: 1) a high-level presentation of the new humanoid DRACO 3 built by Apptronik, where we highlight key design modifications that we have found beneficial to increase the durability and performance of the robot, 2) an extension to our previously presented WBC framework to take into account RCJs by means of internal constraint inclusions, 3) an analysis on the centroidal properties of the DRACO 3 showing how it compares to other known and relatively large humanoids, and 4) a set of preliminary results showing the humanoid performing disturbance rejection and a squatting and swaying task to validate the compliance of our design propositions as well as the integration of our extended WBC to control DRACO 3.

## II. SYSTEM OVERVIEW

In this section, we introduce the overall actuation mechanisms in DRACO 3, placing greater emphasis on the hip and knee mechanisms, which make up the most complex parts of the robot.

### A. Robot Design

DRACO 3 stands 1.35 m tall, weighs 39 kg, and has a range of motion (RoM) similar to that of a human in its actuated degrees of freedom (DoF): neck pitch, 6-DoF arms, and 6-DoF legs. The joint configuration shown in Fig. 1 (a) is used by the low-level controller. DRACO 3's proprioceptive and exteroceptive sensors are illustrated in Fig. 2(c). Its lower-body constitutes the most complex part of the robot as its actuation principle exploits proximal actuation and seeks to achieve a high range of motion.

In general, the robot leverages both off-the-shelf and Apptronik's motors, all of which communicate through EtherCAT by means of Apptronik's motor and sensor boards. The usage of multiple encoders equips the robot with accurate sensing capabilities, even in non-collocated actuated joints, thus resulting in accurate control authority over all joints.

### B. Hip Pitch Design with Cable-Pulley Actuation

The lateral view of the hip is shown in Fig. 2(a). This joint is actuated by means of a rolling contact, with the colored curves delineating the routing of the tensioned cables inside. This design has taken into account several factors: the cable material selection, the termination design, and the sensor placement. Typical materials for cable-driven actuators include stainless steel, Vectran, and Kevlar, among others [5], where it is often desirable to have high stiffness, low creep, and low D-to-d ratio (excluding environmental resistance specifications, e.g., water resistance). Since the bending radii in our design are fairly big in comparison to the required cable diameter, the D-to-d ratio is relaxed, thus favoring stainless steel. On the other hand, terminations for stainless steel tend to be a bit more restrictive. For our design, we have made custom barrel-end cables by silver soldering the cable to customized barrels, and have crimped loop sleeves over vented screws on the other end, as shown in Fig. 2(b). These terminations allow us to overcome some of the drawbacks of using stainless steel. This allows us to reliably pursue this design for the hip with a stainless steel cable of 1.5 mm diameter.

The position of the thigh is then measured with an absolute encoder located at the pivoting joint on the fixed sheave, with its corresponding magnet aligned in the thigh link.

### C. Knee Design with Rolling Contact Joint Mechanism

The knee joint is designed as a rolling contact mechanism to ensure a large RoM. This not only enlarges the workspace of the robot, but also increases its transportability by enabling the legs to be completely foldable. Unlike the hips, in this case, the pivoting point is not fixed but instead occurs at the interface of the rolling contact, and is thus continuously changing. The thigh and shin links are constrained to remain at a fixed distance and to roll without slipping. Thus, as the shin rotates, it simultaneously rotates and rolls along the end of the thigh. Since the radii of the thigh and shin rolling surfaces are the same, the absolute motion of the link w.r.t. each other is twice the rotation of the shin, as shown in Fig.2(d). For instance, $q_{\text{knee}} = 90°$, when $q_{\text{proximal}} = 45°$, due to such combined motion.

Among the different materials for cable-driven actuation mentioned in II-B, a less stiff material was chosen for the knees. This is because, unlike the hip, the knee is kinematically constrained by more elements, thus reducing the overall tension in the cables and allowing us to use a different material instead of stainless steel. The benefit of this is that making the terminations for a set of cables requires no additional machinery, unlike stainless steel. For mechanical simplicity and compactness, an Estar stopper knot was used at the cable termination (see Fig. 2(f)). Although it has been reported in [5] that the common knot terminations in cable-driven actuation show poor performance, in our experience, the Estar stopper knot led to reliable cable termination in our knee design.

Another important design consideration on the knee joint involves proximal actuation to achieve agile and dynamic motions [3]. The quantitative evaluation of DRACO 3's design on proximal actuation will be shown later in section IV-A. A multistage transmission with minimal backlash allows for torques to be transmitted to the knee from the motor located near the hips. The corresponding torques are computed by taking into consideration these transmissions, which results in a high torque density actuation capable of an extensive range of motion.

## III. ROBOT SOFTWARE AND CONTROL

This section describes the software architecture of DRACO 3, our proposed WBC formulation that accounts

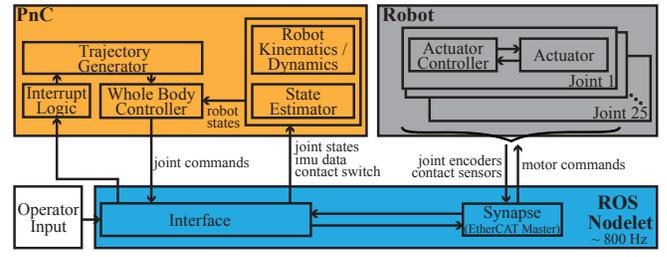

Fig. 3. **DRACO 3 Software Architecture.** Given user input through a ROS service call, *Interrupt Logic* module triggers to generate desired trajectories which are fed into WBC and this leads to compute joint commands in WBC.

for the rolling contact joints, and a direct mapping of our WBC outputs to actuator commands.

### A. Software Architecture

The control architecture of DRACO 3 consists of a decentralized controller, which runs at the actuator level, and a centralized controller, which computes the high-level commands (e.g., WBC). The interconnection between the different components of these controllers is shown in Fig. 3. DRACO 3's control computer runs a ROS nodelet which enables the communication with the robot sensors and actuators. The sensor measurements and desired control commands are then exposed as dedicated ROS messages via the *Synapse* module, and a separate *Interface* module is created to package them for our high-level controller to read and write to. The high-level control actions are computed online through our PnC [10], which is installed as an external library. Given the current sensor measurements and a user command, PnC computes desired joint commands, which are read by the *Interface* and sent to the *Synapse* at the end of each synchronized control loop.

### B. WBC

Here, we describe constraints for the rolling contact joint, a rigid-body dynamic model of constrained robots, and our whole-body control framework called Implicit Hierarchical Whole-Body Control (IHWBC) in [10].

*1) A Model of Rolling Contact Joint:* As shown in Fig. 4, there exist a physical relationship between the knee and distal joints [7] as follows:

$$q_{\text{knee}} = q_{\text{proximal}} + q_{\text{distal}} = \left(1 + \frac{r_{\text{distal}}}{r_{\text{proximal}}}\right) q_{\text{distal}}, \quad (1)$$

where $q_{\text{knee}}$ and $q_{\text{distal}}$ denote the knee and distal joint positions, respectively. In particular, $q_{\text{proximal}} = q_{\text{distal}}$ is always true in the current design. Based on the above relationship, we formulate internal constraints such that

$$\begin{aligned}\dot{\mathbf{x}}_{\text{int}} &= \mathbf{J}_{\text{int}}\dot{\mathbf{q}} = \mathbf{0}, \\ \ddot{\mathbf{x}}_{\text{int}} &= \mathbf{J}_{\text{int}}\ddot{\mathbf{q}} + \dot{\mathbf{J}}_{\text{int}}\dot{\mathbf{q}} = \mathbf{J}_{\text{int}}\ddot{\mathbf{q}} = \mathbf{0}\end{aligned} \quad (2)$$

where $\mathbf{q} \in \mathbb{R}^n$ and $\mathbf{J}_{\text{int}}$ denotes the joint variable and Jacobian for the internal constraints, respectively. In our case,

the Jacobian matrix of the internal constraints becomes a constant matrix:

$$\mathbf{J}_{\text{int}} = \begin{bmatrix} \mathbf{0} & +1 & -1 & 0 & 0 & \mathbf{0} \\ \mathbf{0} & 0 & 0 & +1 & -1 & \mathbf{0} \end{bmatrix}.$$

$\underbrace{\phantom{xxx}}_{\text{floating base}}\ \underbrace{\phantom{xxx}}_{\text{left knee}}\ \underbrace{\phantom{xxx}}_{\text{right knee}}\ \underbrace{\phantom{xxx}}_{\text{the rest joints}}$

We utilize the above internal constraints when generating a model for the robot's constrained dynamics and formulating an optimization problem for our IHWBC.

*2) Dynamics Model:* The rigid-body dynamics of humanoid robots with internal constraints is expressed as follows:

$$\mathbf{A}\ddot{\mathbf{q}} + \mathbf{b} + \mathbf{g} = \mathbf{S}_a^\top \boldsymbol{\tau} + \mathbf{J}_{\text{int}}^\top \mathbf{F}_{\text{int}} + \mathbf{J}_c^\top \mathbf{F}_r, \quad (3)$$

where $\mathbf{A}$, $\mathbf{b}$, $\mathbf{g}$, $\mathbf{S}_a$, and $\boldsymbol{\tau}$ denote the mass/inertia matrix, coriolis/centrifugal force, and gravitational force, actuator selection matrix, and torque command, respectively. $\mathbf{J}_c$, $\mathbf{F}_r$, and $\mathbf{F}_{\text{int}}$ denote the contact Jacobian, reaction forces, and internal forces. Since $\dot{\mathbf{x}}_{\text{int}} = \ddot{\mathbf{x}}_{\text{int}} = \mathbf{0}$, we represent $\mathbf{F}_{\text{int}}$ in terms of $\boldsymbol{\tau}$ and $\mathbf{F}_r$. In turn, we substitute the expressed $\mathbf{F}_{\text{int}}$ into (3) then the following constrained dynamic model is obtained.

$$\mathbf{A}\ddot{\mathbf{q}} + \mathbf{N}_{\text{int}}^\top(\mathbf{b}+\mathbf{g}) + \mathbf{J}_{\text{int}}^\top(\mathbf{J}_{\text{int}}\mathbf{A}\mathbf{J}_{\text{int}}^\top)^\dagger \dot{\mathbf{J}}_{\text{int}}\dot{\mathbf{q}} \\ = (\mathbf{S}_a\mathbf{N}_{\text{int}})^\top \boldsymbol{\tau} + (\mathbf{J}_c\mathbf{N}_{\text{int}})^\top \mathbf{F}_r, \quad (4)$$

where $\mathbf{N}_{\text{int}} = \mathbf{I} - \overline{\mathbf{J}}_{\text{int}}\mathbf{J}_{\text{int}}$ with $\overline{\mathbf{J}}_{\text{int}} = \mathbf{A}^{-1}\mathbf{J}_{\text{int}}^\top(\mathbf{J}_{\text{int}}\mathbf{A}^{-1}\mathbf{J}_{\text{int}}^\top)^\dagger$. Since the Jacobian matrix for the internal constraints is constant, the 3rd term in (4) vanishes. In addition, it is noted that the internal constraint force $\mathbf{F}_{\text{int}}$ does not affect to the torque effects on the first six joints. However, the configuration and velocity of the floating base will be indirectly driven by the actuated part incorporating $\mathbf{F}_{\text{int}}$. In our WBC, We enforce the constraints for the underactuated and actuated dynamics separately so $\mathbf{F}_{\text{int}}$ vanishes in our optimization formulation.

*3) IHWBC:* We employ the IHWBC framework from our previous work [10], which enables smooth task and contact transitions. However, we additionally constrain our solutions to lie within the manifold of the rolling contact mechanism. The resulting optimization algorithm, formulated as a quadratic programming (QP), is written as follows:

$$\min_{\ddot{\mathbf{q}},\mathbf{F}_r}\ \mathcal{J}(\ddot{\mathbf{q}}, \mathbf{F}_r) \quad (5a)$$

$$\text{s.t.}\ \mathbf{S}_f\left(\mathbf{A}\ddot{\mathbf{q}} + \mathbf{b} + \mathbf{g} - \mathbf{J}_c^\top \mathbf{F}_r\right) = \mathbf{0}, \quad (5b)$$

$$\mathbf{J}_{\text{int}}\ddot{\mathbf{q}} + \dot{\mathbf{J}}_{\text{int}}\dot{\mathbf{q}} = \mathbf{0}, \quad (5c)$$

$$\mathbf{U}\mathbf{F}_r \geq \mathbf{0}, \quad \mathbf{S}_r\mathbf{F}_r \leq \mathbf{F}_r^{\max}, \quad (5d)$$

$$\ddot{\mathbf{q}}_{\min} \leq \ddot{\mathbf{q}} \leq \ddot{\mathbf{q}}_{\max}, \quad (5e)$$

$$\boldsymbol{\tau}_{\min} \leq \overline{\mathbf{S}_a\mathbf{N}_{\text{int}}}^\top \left(\mathbf{A}\ddot{\mathbf{q}} + \mathbf{N}_{\text{int}}^\top(\mathbf{b}+\mathbf{g})\right. \\ \left. - (\mathbf{J}_c\mathbf{N}_{\text{int}})^\top \mathbf{F}_r\right) \leq \boldsymbol{\tau}_{\max} \quad (5f)$$

and

$$\mathcal{J}(\ddot{\mathbf{q}}, \mathbf{F}_r) = \sum_{i=1}^n \left\|\mathbf{J}_i\ddot{\mathbf{q}} + \dot{\mathbf{J}}_i\dot{\mathbf{q}} - \ddot{\mathbf{x}}_i^d\right\|_{\mathbf{W}_i}^2 + \lambda_q \|\ddot{\mathbf{q}}\|^2 \\ + \left\|\mathbf{F}_r^d - \mathbf{F}_r\right\|_{\mathbf{W}_{F_r}}^2 + \lambda_{F_r} \|\mathbf{F}_r\|^2 \quad (6)$$

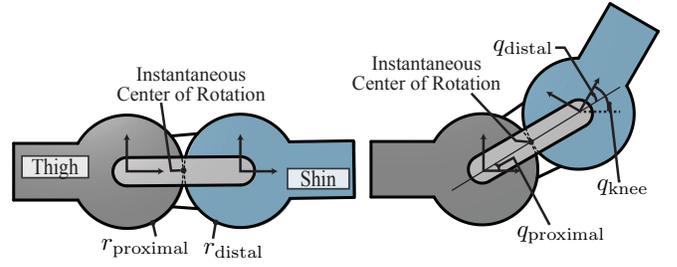

Fig. 4. **Schematics of the rolling contact joint on knee.** Zero configuration (left) and after some angular displacement (right).

where $\mathbf{W}_i$ and $\mathbf{W}_{F_r}$ denote the weighting matrices for the $i$-th task and the reaction force. In addition, $\lambda_q$ and $\lambda_{F_r}$ are the weighting values for the norms of the decision variables in the cost function.

The constraints in this QP are the floating base dynamics (5b), the rolling contact motion constraint (5c), the unilateral constraint, cone wrench constraint, and reaction force constraint (5d), joint acceleration constraint (5e), and torque constraint (5f). The running cost penalizes for errors on task and reaction force trajectory tracking, and includes regularization terms on the joint accelerations and reaction forces. Given the desired task specifications $\ddot{\mathbf{x}}_i^d$ and desired reaction forces $\mathbf{F}_r^d$, the optimal joint accelerations $\ddot{\mathbf{q}}^\star$ and reaction forces $\mathbf{F}_r^\star$ are computed. The desired joint torques $\boldsymbol{\tau}^\star$ are subsequently obtained using inverse dynamics while the desired joint velocities and positions are computed by integrating $\ddot{\mathbf{q}}^\star$, as in [10]. Note that, in order to take into account the rolling contact joint, we have introduced internal constraints (5c) and modified the actuator saturation constraints (5f). For more details to other than these changes, the reader is referred to [10].

*4) Torque Limits:* The actuator saturation constraints are derived by rearranging Eq. (4) with the dynamically consistent inverse of $(\mathbf{S}_a\mathbf{N}_{\text{int}})^\top$, as denoted by the over-bar operator in (5f). We truncate the matrix $\mathbf{S}_a\mathbf{N}_{\text{int}}$ by removing the columns corresponding to the floating base for computational efficiency. Note that Eq. 5f is only valid when $\overline{\mathbf{S}_a\mathbf{N}_{\text{int}}}\mathbf{S}_a\mathbf{N}_{\text{int}} = \mathbf{N}_{\text{int}}$ (without the floating base components), which is satisfied in DRACO 3 at all configurations.

### C. Actuation Mapping

Due to the non-collocated actuation at the hip and knee joints, the torques computed by the WBC presented in the previous section need to be mapped to the actual motor locations for control. These desired torques are applied in a feed-forward fashion in the low-level joint controller. Here, we present the aforementioned mappings.

*1) Hip Pitch Joint:* The hip pitch torque command computed by the WBC is expressed at the center of the fixed sheave shown in Fig. 2(a), which is where the revolute joint is defined in our model. This torque is mapped from the hip joint axis to the motor axis through an effective transmission ratio of $r_{\text{fix}}/r_{\text{rot}}$, where $r_{\text{fix}}$ and $r_{\text{rot}}$ are the radius of the fixed and rotating sheaves, respectively.

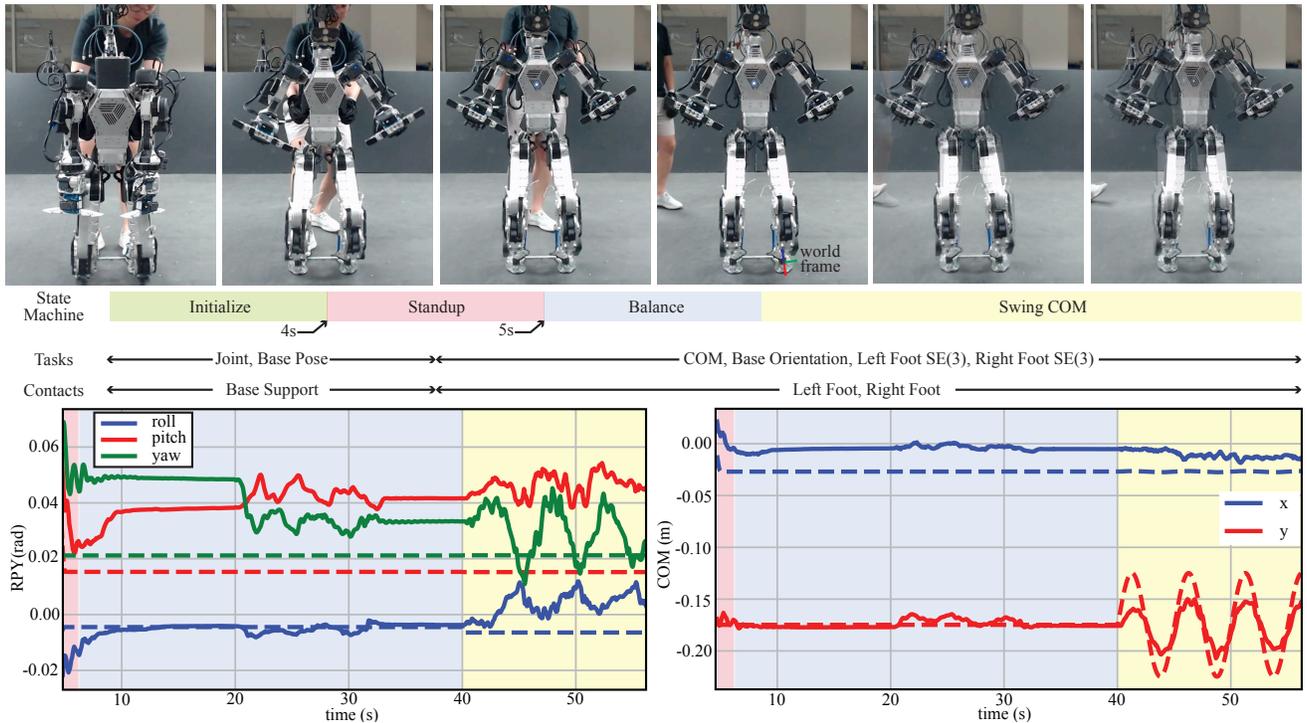

Fig. 5. **Disturbance rejection and COM swing test.** Snapshots of the experiment as the robot progresses through the different states in the state machine (top). Desired (dashed lines) and actual (solid lines) trajectories of the roll-pitch-yaw of the torso (bottom left). Desired (dashed lines) and actual (solid lines) $x$- and $y$-position of the CoM w.r.t. world frame as shown in top snapshots (bottom right). In both bottom figures, the disturbances between the 20 and 30 seconds correspond to three different manual pushes performed on the robot prior to switching to the "SwingCOM" state.

*2) Knee Pitch Joint:* The knee torque command computed by the WBC is defined on the knee distal joint axis (i.e., $q_{\text{distal}}$ in Fig. 4). Thus, we first map it to the ICR of the rolling contact, then use the transmission ratios to map it to the motor location. Considering the transmission relation from Eq. (1), we obtain the effective torque exerted on the ICR axis (which, in our case, is half the torque applied on the knee distal joint axis). This torque is subsequently transformed into the motor axis via the corresponding transmission ratios, as in [8].

## IV. RESULTS AND DISCUSSIONS

### A. Design Evaluation

We seek to quantify the mechanical improvements gained from the proximal actuation design on DRACO 3. At the same time, we illustrate how DRACO 3 compares against other adult-size humanoid robots. As a metric to quantify the inertial contribution of limb motions we consider the Centroidal Inertia Isotropy (CII) metric proposed in [3], which evaluates a system's proximodistal mass distribution for a nominal and current joint configuration pair. In particular, for any set of test configurations, $\mathbf{Q}$, the CII is defined as

$$\text{CII}(\mathbf{q}, \mathbf{q_0}) := \det(\mathbf{I}_G^{-1}(\mathbf{q})\mathbf{I}_G(\mathbf{q_0}) - \mathbf{1}_3),$$

where $\mathbf{q}, \mathbf{q_0} \in \mathbf{Q}$, $\mathbf{I}_G \in \mathbb{R}^{3\times 3}$ are the configuration, nominal configuration, and rotational centroidal composite rigid body inertia matrix [15], respectively.

When generating the CII values, we chose the nominal configurations such that the robots are in the upright position with their arms fully extended to the sides and their knees bent by 90 degrees. We then computed their CII values by generating one-step motions in Cartesian space, where the step length of each robot is normalized based on the robot's height for fair comparison across robots. We then obtained the set of test configurations by using inverse kinematics, resulting in a set of 3000 configurations per robot, and present their associated CII values in Fig. 6. Note that along with DRACO 3 (proximal), we have considered as well the case of DRACO 3 (collocated), which we have created by placing the knee motor and its transmission (corresponding to 2kg) at the knee instead of its current location. This lead to two main results. First, we notice that the proximal design of DRACO 3 reduces the effect of leg inertia by a 36%, as obtained by the range of CII, which goes from 0.00033 to 0.00021. Second, we see that although the leg inertia is reduced, it remains comparable to other conventional electric motor robots, such as HUBO 2. This suggests that conventional reduced-order models might not readily work on this type of robot without further taking into account limb inertia motions.

### B. Hardware Experiments

DRACO 3's control PC is located off-board and runs on Ubuntu 18.04 LTS patched with the RT-Preempt kernel, which enables real-time control performance. Each actuated joint has an embedded control board for motor control and communicate with the control computer via EtherCAT. The robot is powered by an external power supply.

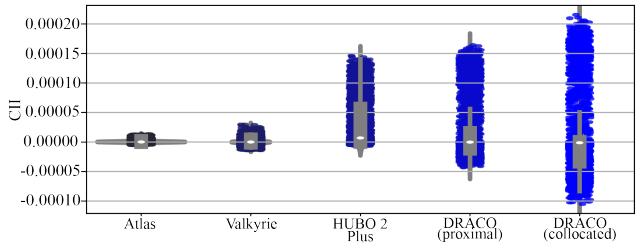
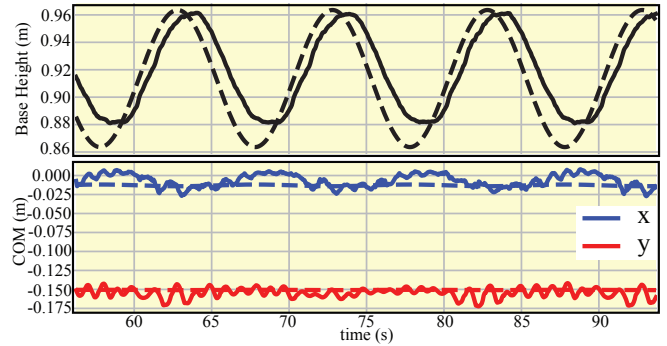

Fig. 6. **CII comparisons.** Distributions of CII computed for test configurations corresponding to a single-step walking trajectory for the adult-size humanoid robots in [2], [16], [17] and DRACO 3 presented here. A large absolute value of CII means significant inertia changes compared to the nominal pose. For DRACO 3, the swing foot target (stepping) location is varied within [0.1, 0.2] meters forward and [-0.1, 0.1] meters laterally with a swing height of 0.05m. 30 data points are sampled along each swing trajectory. The base trajectory was sampled in the middle of the stance and swing foot.

Fig. 7. **Squatting Test.** Base trajectory of DRACO 3 performing a squatting test showing its performance in the z-direction (top) and the $x$- and $y$-directions. The reference and actual measurements are shown in dashed and solid lines, respectively.

We perform two main hardware experiments to validate our proposed WBC implementation for RCJs on DRACO 3. Both of these are carried out using the same controller architecture with the same desired tasks, specifying an: orientation task, feet position and orientation task, balance task through control of the instantaneous capture point, upper body joint positions task, and regularization terms on ground reaction forces. Notice that unlike our previous formulation in [10], here we have implemented the foot contact task by fixing the desired foot position to a single value with zero desired velocity, which resulted in more assertive and less oscillatory corrections by the controller. In addition, it was our experience that controlling the base frame instead of the CoM frame resulted in better performance and thus, present our results pursuing this approach. The states from the state machine used for these experiments are shown in Fig. 5.

*1) Disturbance Rejection:* In this experiment, we waited for the robot to reach the "Balance" state and perturbed it by manually performing first a constant push and then an impulsive push, both in the lateral direction. The constant push moved the estimated center of mass of the robot by roughly 3 cm from its desired configuration as the robot pushes against the disturbance to remain in balance. After releasing the robot from the disturbance, the robot bounced back in direction opposite to the push and managed to stabilize again. The impulsive push was applied in a similar fashion but for a shorter time duration, to which the robot again reacted to stabilize.

*2) CoM Tracking:* The purpose of this experiment is to show the ability of our WBC and overall control framework to track a non-static reference while fulfilling other desired tasks. The task priorities were given as follows: feet hierarchy set to 40, ICP hierarchy set to 20, all upper joint positions set to 20. We again waited for the robot to reach the "Balance" state, then we manually triggered the robot into the "SwingCOM" state. The desired trajectory is a sinusoidal reference, in this case given in the world frame, which is coincident and aligned with the left foot, as shown in the fourth snapshot in Fig. 5. The left graph shows the performance of the tracking of the base orientation task,
showing that the roll is tracked almost perfectly throughout this test, while the pitch and yaw remain within 1 deg of the desired pose, even during the swing motion. On the other hand, the graph on the right shows the performance of the tracking of the CoM during the "Balance" and "SwingCOM" states. Although the controller is able to track the reference signal with minimal lag, it remains 2 cm short from tracking the 5 cm amplitude. This is due to our current low control bandwidth in the ankle roll actuators, which we plan to further improve as future work. A similar experiment is conducted with DRACO 3 squatting, shown in Fig. 7. It can be seen that in this case the base trajectory is tracked more accurately in all directions, indicating that our WBC is accurately producing the required torques meeting the specified tasks.

Our future work will include further experiments carrying out more complex tasks, including but not limited to walking and performing legged manipulation tasks.

## V. CONCLUSION

This work briefly introduced the new humanoid DRACO 3 developed by Apptronik and further improved for extended durability by the authors. One of the key components that in our experience was key to this improvement was the material selection in the cable-driven systems. We leveraged a combination of stainless steel and fiber cables in different parts of the robot and have found these to be more efficient in DRACO 3's actuation mechanisms. We also quantify through the CII metric a reduced effect of leg inertia by using a proximal actuation. More importantly, we present a WBC that incorporates rolling contact mechanisms by considering it as an internal constraint, show how this is easily integrated into our open-source planning and control framework, and then deployed and tested on DRACO 3 while performing distrubance rejection and CoM tracking tasks.

## ACKNOWLEDGMENT

The authors would like to thank the members of the Human Centered Robotics Laboratory at The University of Texas at Austin and Apptronik Systems, Inc. for their great help and support.